\documentclass[conference]{IEEEtran}
\IEEEoverridecommandlockouts
\usepackage{cite}
\usepackage{amsmath,amssymb,amsfonts}
\usepackage{algorithm}
\usepackage{algorithmicx}
\usepackage{algpseudocode}
\usepackage{graphicx}
\usepackage{textcomp}
\usepackage{xcolor}
\usepackage{balance}
\usepackage{booktabs}
\usepackage{multirow}
\emergencystretch=\maxdimen
\def\BibTeX{{\rm B\kern-.05em{\sc i\kern-.025em b}\kern-.08em
    T\kern-.1667em\lower.7ex\hbox{E}\kern-.125emX}}

\begin{document}

\title{FedADP: Unified Model Aggregation for Federated Learning with Heterogeneous Model Architectures\\
}
\DeclareRobustCommand*{\IEEEauthorrefmark}[1]{%
  \raisebox{0pt}[0pt][0pt]{\textsuperscript{\footnotesize #1}}%
}
\author{
    \IEEEauthorblockN{
        Jiacheng Wang\IEEEauthorrefmark{1}, 
        Hongtao Lv\IEEEauthorrefmark{1} and 
        Lei Liu\IEEEauthorrefmark{1, 2}\IEEEauthorrefmark{*}\thanks{* Corresponding author: Lei Liu.}
    }
    \IEEEauthorblockA{
        \IEEEauthorrefmark{1}School of Software, Shandong University, Jinan 250000, China\\
        \IEEEauthorrefmark{2}Shandong Research Institute of Industrial Technology, Jinan 250000, China\\
        Email: 
            jiachengwang@mail.sdu.edu.cn,
            lht@sdu.edu.cn, 
            l.liu@sdu.edu.cn
    }
}

\maketitle

\begin{abstract}
Traditional Federated Learning (FL) faces significant challenges in terms of efficiency and accuracy, particularly in heterogeneous environments where clients employ diverse model architectures and have varying computational resources. Such heterogeneity complicates the aggregation process, leading to performance bottlenecks and reduced model generalizability. To address these issues, we propose FedADP, a federated learning framework designed to adapt to client heterogeneity by dynamically adjusting model architectures during aggregation. FedADP enables effective collaboration among clients with differing capabilities, maximizing resource utilization and ensuring model quality. Our experimental results demonstrate that FedADP significantly outperforms existing methods, such as FlexiFed, achieving an accuracy improvement of up to $23.30\%$, thereby enhancing model adaptability and training efficiency in heterogeneous real-world settings.

\end{abstract}

\begin{IEEEkeywords}
distributed computing, federated learning, architecture heterogeneity.
\end{IEEEkeywords}

\section{Introduction}
Federated learning (FL) is a decentralized machine learning approach that enables multiple clients to collaboratively train a shared model while keeping their data localized, thereby preserving privacy and reducing data transfer. Traditional FL faces multiple challenges, primarily due to the assumption that all clients use models with identical architectures~\cite{li2020federatedz}. In reality, differences in hardware capabilities, such as computational power, memory size, and bandwidth—often require clients to use models with different architectures to accommodate resource limitations~\cite{wang2019adaptive, fallah2020personalized2}. This discrepancy becomes problematic when devices with limited computational power hold crucial data, creating bottlenecks in the training process~\cite{yu2020deep, diao2020heterofl, wu2023efficient}. Excluding these devices wastes valuable data and resources, ultimately compromising the accuracy and generalizability of the model~\cite{wang2019adaptive, imteaj2022federated}. 

One direction of personalized federated learning (PFL) has emerged to address these challenges, enabling clients with different model architectures to collaborate in training~\cite{smith2017federated}. FlexiFed~\cite{wang2023flexifed} is an example of this approach, allowing clients to utilize varying model structures while maintaining global knowledge through sharing. However, methods like FlexiFed aggregate only the common layers of different models, discarding unique layers even when differences are minimal, leading to wasted computational resources and reduced accuracy. These limitations highlight the urgent need for federated learning methods that can adapt effectively to heterogeneous model structures.

To address the heterogeneity of the model in PFL, we propose FedADP, focusing on a holistic model approach. FedADP can not only handle more common heterogeneous network structures but also ensure that even devices with weak computational power can make effective contributions to the training of the global model while protecting data privacy. FedADP maximizes resource utilization and adapts to real-world environments with diverse computational capabilities. In this framework, different edge devices use models such as VGG models~\cite{simonyan2014very} for training. These models dynamically adjust their structures to align with the global model during aggregation, and the global model adapts to the edge devices during distribution. The main contributions of this paper include:

\begin{itemize}
\item This work develops a FedADP approach, which can adaptively train all clients with models of different structures and aggregate them into the same global model.
\item The comparisons of the results between FedADP and exsiting methods, i.e., FlexiFed~\cite{wang2023flexifed} and Clustered-FL~\cite{sattler2020clustered}, suggest the significant improvement of accuracy by up to $23.30\%$ and $46.25\%$.
\end{itemize}

\begin{figure}
    \centering
    \includegraphics[width=1\linewidth]{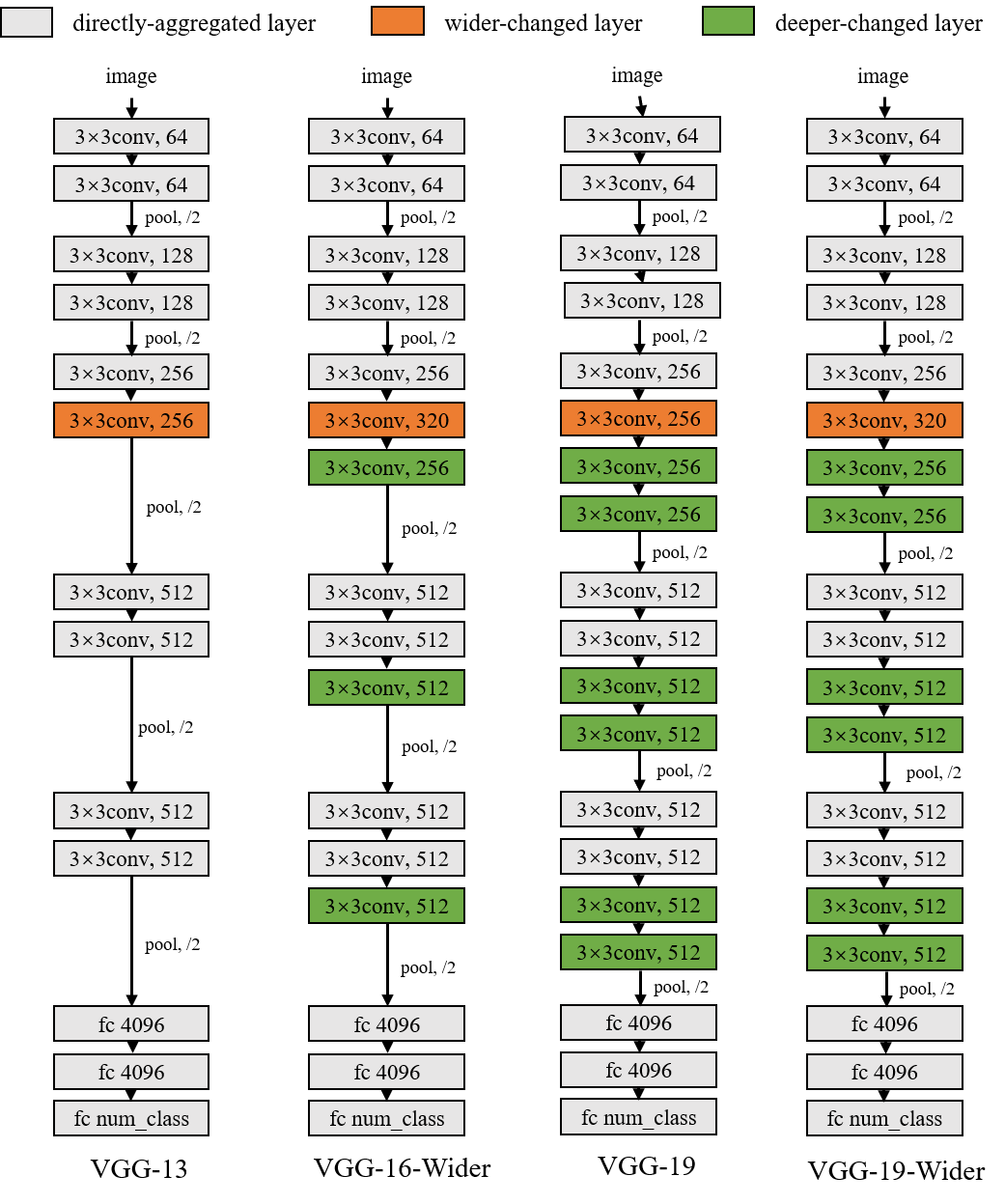}
    \caption{Some variations of the VGG model.}
    \label{fig1}
\end{figure}

\section{Related Work}
\subsection{Personalized Federated Learning}
Personalized federated learning (PFL) enables clients to tailor model updates to local data, improving relevance and performance while preserving privacy. One of the directions of PFL is to enable clients with different model structures to cooperate with each other and adapt to the actual situation to the greatest extent possible. Neural Architecture Search (NAS)~\cite{zoph2016neural} is employed in PFL to optimize personalized models by automatically selecting the most suitable architecture based on the unique data and computational constraints of each client. This is particularly important in federated settings, where clients may differ significantly in computational power and structure characteristics. SPIDER~\cite{mushtaq2021spider}, for instance, dynamically searches for architectures that balance accuracy and efficiency, tailored to each client's specific needs. Another approach, FedMN~\cite{wang2022personalized}, uses a pool of submodels to construct client-specific architectures adaptively, ensuring effective personalization while addressing computational limitations.

In addition, many methods with other strategies, such as Ditto~\cite{Li2020DittoFA}, FedMoE~\cite{mei2024fedmoe}, in which each client participates in global model training while maintaining and updating its personalized local model. The above solutions provide various approaches for PFL with heterogeneous model architectures, but their core strategy remains to aggregate only the common layers while discarding the different ones. In contrast, our proposed FedADP adjusts model architectures to aggregate all potential clients, addressing the data loss and accuracy caused by such strategies.

\subsection{Heterogeneous Network Structure}
The capabilities of heterogeneous devices in FL pose challenges due to the varying update rates and resource constraints, affecting convergence and global performance~\cite{li2020federated, li2022federated, mohammadabadi2023speed, zhang2022fedada}. Methods like Net2Net~\cite{chen2015net2net} and ModelKeeper~\cite{lai2023modelkeeper} facilitated efficient knowledge transfer, reducing retraining time, and our work builds on these approaches to address the complexities of such environments. 
Several recent studies have attempted to solve the challenge of heterogeneous model architectures in federated learning (FL). Hetefedrec~\cite{yuan2024hetefedrec} proposed a federated recommender system to handle the heterogeneity of the model between clients. HDHRFL~\cite{jiang2024hdhrfl} introduced a hierarchical framework for robust FL in dual-heterogeneous and noisy client environments. AdapterFL~\cite{liu2023adapterfl} developed an adaptive approach for FL in resource-constrained mobile systems with heterogeneous models.  FIARSE~\cite{wu2024fiarse} used importance-aware submodel extraction to facilitate model-heterogeneous FL. VFedMH~\cite{wang2023vfedmh} proposed a vertical federated learning framework to train multi-party heterogeneous models. Most of these methods focus on the homogeneous components within heterogeneous models. Specifically, they aim to identify or even construct common parts across different model architectures and leverage them effectively. However, knowledge outside the common structural components often remains unused, leading to considerable waste of both knowledge and computational resources. In contrast, FedADP abandons the attempt at local structural similarity and instead directly modifies heterogeneous models to a unified structure for aggregation. This approach maximizes the use of computational power and knowledge from each client while maintaining applicability even in scenarios with high degrees of heterogeneity.

\section{FedADP}
This section mainly describes the design of FedADP. Prior to introducing FedADP, we shall briefly present traditional FL first. Given $K$ clients, the global model update of traditional FL can be described by the following equation:
\begin{equation}
\mathbf{\omega}^{t+1} = \sum_{k=1}^{K} {W_k} \mathbf{\omega}_k^t, 
\end{equation}
and
\begin{equation}
{W_k} =  \frac{n_k}{n} 
\end{equation}
where ${\omega}^{t+1}$ is the updated global model at round $t+1$, ${\omega}_k^t$ is the local model from client $k$ at round $t$, ${W_k}$ denotes the weight of each client in aggregation, ${n_k}$ denotes the number of data samples at client $k$, and \( n = \sum_{k=1}^{K} n_k \) is the total number of data samples across all clients. Clients perform several rounds of local updates, often using gradient-based methods such as stochastic gradient descent (SGD)~\cite{robbins1951stochastic}. The local model update on both traditional FL and FedADP at client $k$ can be expressed as:
\begin{equation}
\mathbf{\omega}_k^{t+1} = \mathbf{\omega}_k^t - \eta \nabla F_k(\mathbf{\omega}_k^t)
\end{equation}
where \( \mathbf{\omega}_k^{t+1} \) represents the updated model at client \( k \) after local training in round \( t+1 \), \( \mathbf{\omega}_k^t \) is the current model at client \( k \) in round \( t \), \( \eta \) is the learning rate, and \( \nabla F_k(\mathbf{\omega}_k^t) \) denotes the gradient of the local loss function \( F_k \) at client \( k \).

\subsection{FedADP Framework}
Compared with traditional FL, FedADP allows learning between different network structures. Although some traditional PFL methods also allow different model structures, they essentially still pursue the maximum search and utilization of the same local structure~\cite{kulkarni2020survey}, which limits the application scope of PFL. To mitigate the waste caused by excessive focus on local structural similarity and to more effectively utilize the computational power and data resources of each client. Different with most current personalized federated learning solutions, FedADP enables all clients to directly participate in the training and aggregation process by modifying the model structure, eliminating the need to identify identical substructures. FedADP introduces a novel approach to address model heterogeneity by transforming client models with differing architectures into a unified structure prior to aggregation. This transformation allows models with different complexities to be effectively combined, ensuring a consistent aggregation process. At the start of a new training round, the aggregated models are reverted to their original configurations and redistributed back to the respective clients. This approach maintains consistency during aggregation while preserving the unique characteristics and adaptability of each model, enabling efficient collaboration without sacrificing local model optimization. The difference between traditional PFL and FedADP in the workflow is shown in Fig. 2. Algorithm 1 shows the basic steps of FedADP, $NetChange (a, b)$ denotes the application which modifies $a$ by adding or pruning to ensure it conforms to the same network structure as $b$, while preserving the original data. The detailed steps and codes for NetChange will be explained below. 

\begin{itemize}
\item Step 1: Initialize the global model \(\omega^0\).
\item Step 2: The global model performs To-Shallower and To-Narrower of NetChange and is distributed to each client $k$.
\item Step 3: Each client $k$ is trained locally to obtain the local model \(\omega_k\).
\item Step 4: Perform To-Deeper and To-Wider of NetChange on each \(\omega_k\) to keep it in the same structure as the global model and send it to the server.
\item Step 5: Aggregate each \(\omega_k\) using FedAvg to obtain the updated global model \(\omega^t\).
\item Step 6: Repeat Step 2-Step 5 until convergence.
\end{itemize}

\begin{algorithm}[t]
\caption{FedADP}
\label{algorithm_1}
\textbf{Input}: Client set $\mathbb{U}$, learning rate $lr$, local epoch $E$ number of rounds $R$, current round $t$, each weight $W_k$ of client $k$ \\
\textbf{Output}: Global model \(\omega^R\)
\begin{algorithmic}[1]
\State $t = 0$
\State Initialize global model \(\omega^0\) at round $0$
\While{$t < R$}
    \State Select a set of clients $\mathbb{C}^t$ from $\mathbb{U}^t$
    \For{each model $\omega_k$ of client $k$ in $\mathbb{C}^t$}
        \State $\omega_k \gets \text{NetChange}(\omega^t, \omega_k)$          
    \EndFor

    \For{each model $\omega_k$ of client $k$ in $\mathbb{C}^t$}
        \State Local training for $\omega_k$  
        
        \State $ \omega_k \gets \text{NetChange}(\omega_k, \omega^t)$          
    \EndFor
    \State $\omega^t \gets \sum_{k=1}^{K} {W_k} \mathbf{\omega}_k$ 

    \State $t \gets t + 1$
\EndWhile
\end{algorithmic} 
\end{algorithm}

\subsection{NetChange}
NetChange is one of the core components of FedADP, designed to adjust the model structure by transforming it into deeper, wider, or combined configurations, and conversely, it can also alter the structure to be shallower and narrower. It allows clients of any model structure to fully participate in the aggregation. Our work of NetChange extends the Net2Net~\cite{chen2015net2net} method, known for its ability to modify model structures, which serves as a foundational approach. As an extension of this work, NetChange enables not only the deepening and widening of models but also the capability to make them shallower and narrower. It bridges the gap between local models and the global model, aligning all models to have the same structure. This process ensures maximum information sharing between all clients. 
Before aggregating the models from different clients, the system first constructs a global model by taking the union of the structures of all the client models. For example, as illustrated in Fig. 1, VGG-13, VGG-16-Wider, VGG-19, and VGG-19-Wider are examined. Among these, VGG-16-Wider is derived from the conventional VGG-16 by widening one of its layers, with VGG-19 following a similar modification. The layers that require modification are highlighted in the accompanying diagram with different colors. The global model would be set to VGG-19-Wider. 

\begin{figure}
    \centering
    \includegraphics[width=1\linewidth]{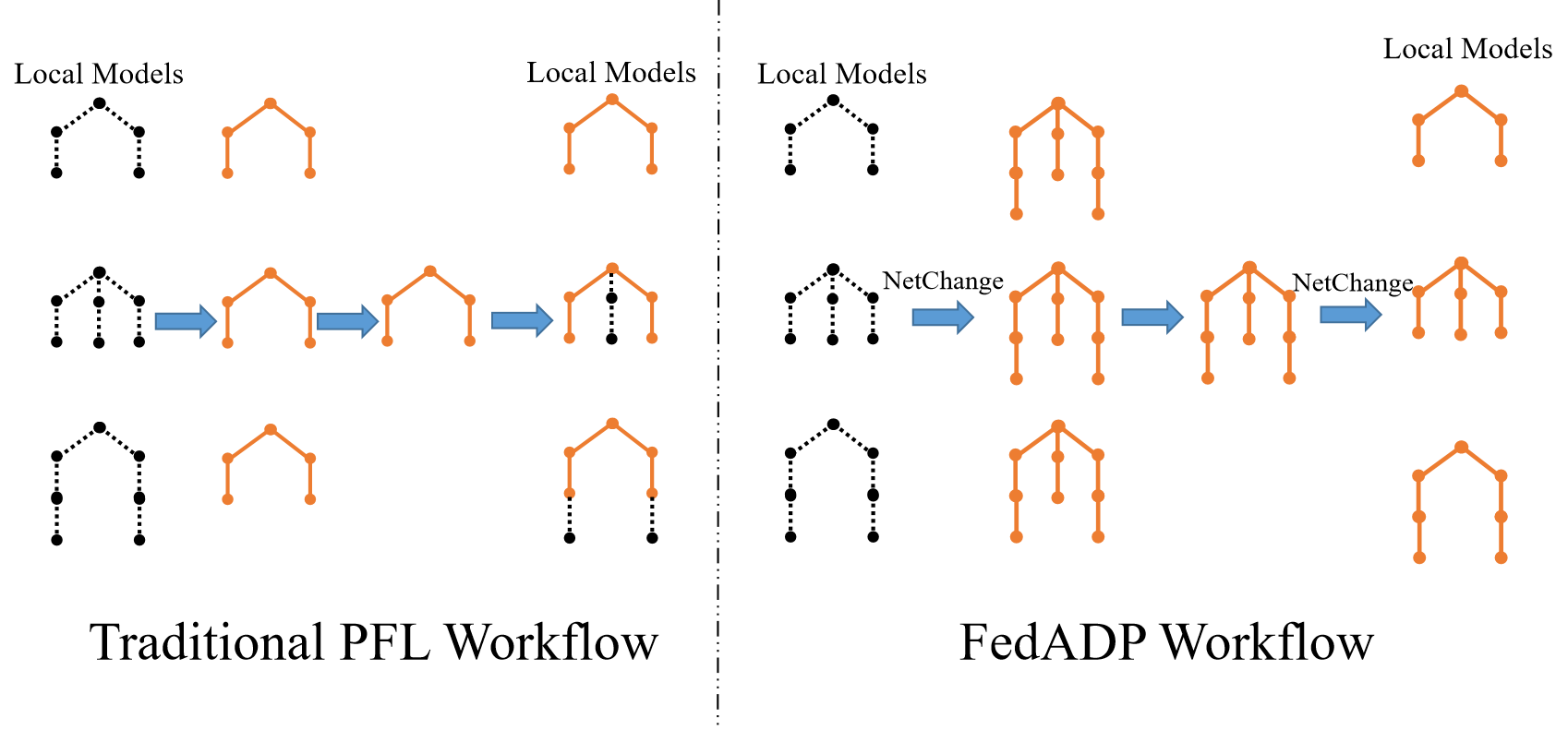}
    \caption{Difference between traditional PFL and FedADP in the workflow}
    \label{fig2}
\end{figure}

\subsubsection{To-Wider and To-Deeper}
After determining the global model, NetChange examines the differences between each local model and the global model. Based on these differences, it decides whether to apply the To-Wider, To-Deeper strategies, or both, to adjust and expand the model structures. This process ultimately aligns the structures of all models with the global model.
During the widening process, additional neurons are created by duplicating existing neurons and their incoming connections, ensuring the new neurons perform the same function as the originals. The weights of these duplicate connections are adjusted so that the output of the expanded layer remains unchanged. When discrepancies exist between the local model and the global model at the layer level, NetChange identifies a layer within the local model that shares the same structure as the missing layer. It then creates a new layer with an identical structure and fills it with specific data. Typically, the diagonal elements are initialized with a value of 1, while all other positions are filled with $0$. Algorithm 2 is the To-Wider code of the NetChange, where one layer $r$ of a client model $\omega_k$ is widened.

\begin{algorithm}[t]
\caption{To-Wider of the NetChange}
\label{alg:Wider}
\textbf{Input}: Client model $\omega_k$, target layer $r$, existing neurons set $\mathbb{C}$ and target neurons set $\mathbb{D}$ \\
\textbf{Output}: Changed layer $r$ 
\begin{algorithmic}[1] %[1] enables line numbers=1
\Function{To-Wider}{$r$, $\mathbb{C}$, $\mathbb{D}$}
            \For{$i$ in $\mathbb{C}$}
                \State $\mathbb{M}_i \gets \{i\}$
            \EndFor
            \For{$i$ in $\mathbb{D} \setminus \mathbb{C}$}
                \State Randomly select neuron $j$ from neurons in $\mathbb{C}$
                \State Set the value $v_i$ of $i$ as the value $v_j$ of $j$ 
                \State Create a new neuron $i$ with $v_i$ to the layer $r$
                \State $\mathbb{M}_i \gets \mathbb{M}_i \cup \{j\}$
            \EndFor
            \For{$i$ in $\mathbb{C}$} 
                \For{$j$ in $\mathbb{M}_i$}
                    \State $v_j \gets v_j/\vert\mathbb{M}_i\vert$
                \EndFor
            \EndFor
        \State \Return  Changed layer $r$ 
\EndFunction 
\end{algorithmic}
\end{algorithm}

\subsubsection{To-Narrower and To-Shallower}
During the model distribution phase, the server customizes the global model by trimming it according to the specific structure of each client, ensuring that the model aligns perfectly with the client's architecture before distribution. This tailored adjustment enables clients to use models that match their computational capabilities and resource constraints. When performing the narrowing operation, excess neurons are removed, and their associated weights are evenly redistributed among the remaining neurons. This redistribution is followed by adjustments to the input connection weights to ensure that the model retains its original functionality and performance. The shallowing operation is simpler and involves the removal of unnecessary layers to match the client’s model structure, thus reducing the complexity of the model while maintaining its effectiveness. Algorithm 3 presents the implementation details of the To-Narrower operation in the NetChange process, providing a systematic approach for reducing the model size to better suit client requirements.

\begin{algorithm}
\caption{To-Narrower of the NetChange}
\label{alg:Narrower}
\textbf{Input}: Client model $\omega_k$, target layer set $\mathbb{T}$, target width $N_{tar}$ \\
\textbf{Output}: Changed model ${\omega_k}\prime$
\begin{algorithmic}[1] %[1] enables line numbers=1
\Function{To-Narrower}{Client model $\omega_k$, target layer set $\mathbb{T}$, new width $N$}  
        \For{each layer $r$ in $\mathbb{T}$}  
            \State $s$ $\gets$ the sum of neurons value after $N_{tar}$ in $r$
            \State Delete the neurons after $N_{tar}$ in $r$
            \State Add $s/N_{tar}$ to each remaining neuron
        \EndFor          
        \State \Return{Changed model ${\omega_k}\prime$}  
\EndFunction 
\end{algorithmic}
\end{algorithm}

\begin{table*}[ht]
\caption{Experimental results on different datasets comparing FedADP, FlexiFed, Clustered-FL, and Standalone under different conditions.}
\centering
\begin{tabular}{c|c c c c}
\toprule
\textbf{} & \textbf{FedADP} & \textbf{FlexiFed} & \textbf{Clustered-FL} & \textbf{Standalone} \\ \midrule
MNIST     & \rule{0pt}{3.5ex} \textbf{0.9834} $\pm$ \textbf{0.0011} & 0.9786 $\pm$ 0.0007 & 0.9697 $\pm$ 0.0023 & 0.9645 $\pm$ 0.0010 \\ 
F-MNIST   & \rule{0pt}{3.5ex} \textbf{0.9053} $\pm$ \textbf{0.0019} & 0.8929 $\pm$ 0.0015 & 0.8860 $\pm$ 0.0069 & 0.8589 $\pm$ 0.0031 \\
CIFAR-10  & \rule{0pt}{3.5ex} \textbf{0.7775} $\pm$ \textbf{0.0035} & 0.7470 $\pm$ 0.0035 & 0.7098 $\pm$ 0.0043 & 0.6126 $\pm$ 0.0013 \\ 
CIFAR-100 & \rule{0pt}{3.5ex} \textbf{0.3630} $\pm$ \textbf{0.0048} & 0.2944 $\pm$ 0.0029 & 0.2482 $\pm$ 0.0026 & 0.1689 $\pm$ 0.0015 \\ \bottomrule
\end{tabular}
\end{table*}

\begin{figure}
    \centering
    \includegraphics[width=1\linewidth]{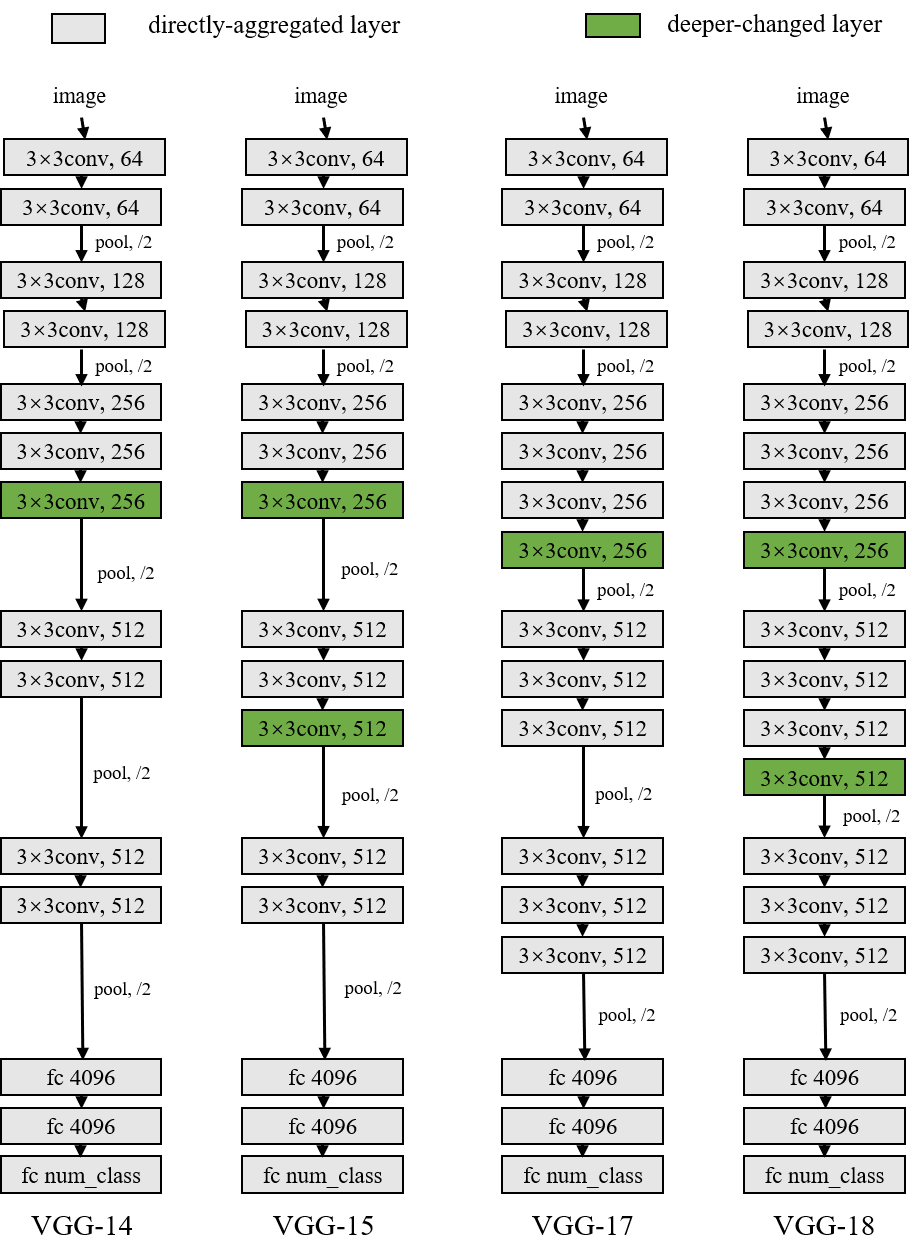}
    \caption{Other forms of the VGG model: VGG-14, VGG-15, VGG-17, and VGG-18.}
    \label{fig3}
\end{figure}

\section{EVALUATION}
\begin{figure*}
    \centering
    \includegraphics[width=1\linewidth]{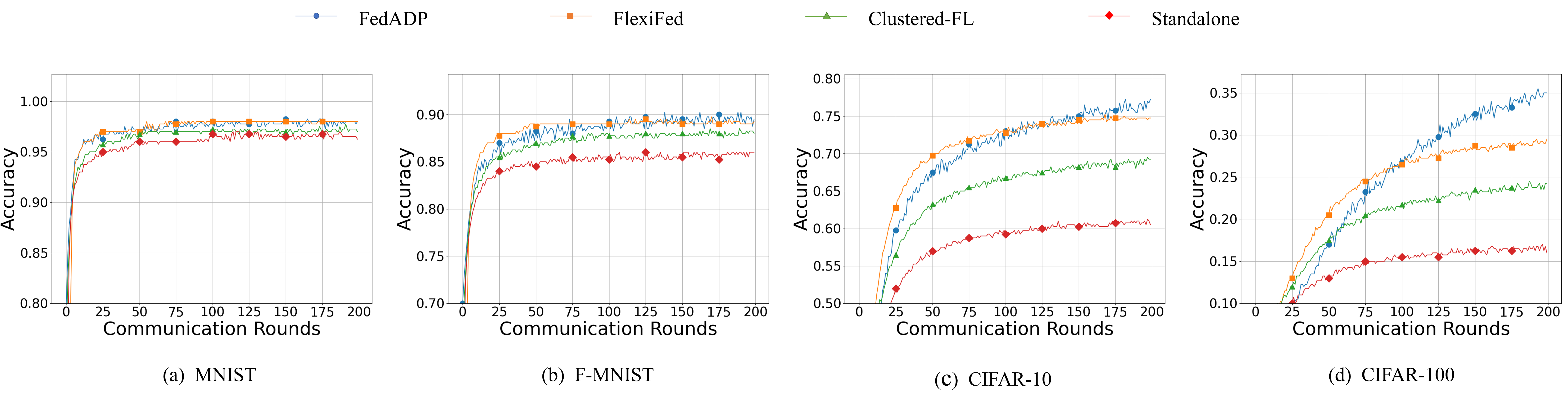}
    \caption{The experimental results compare the performance of FedADP, FlexiFed, Cluster-FL, and Standalone on the MNIST, F-MNIST, CIFAR-10, and CIFAR-100 datasets. }
    \label{fig4}
\end{figure*}

To validate the performance of our work in FedADP, particularly its ability to maintain high accuracy in scenarios with high heterogeneity, we perform a series of experiments using various VGG model~\cite{simonyan2014very} architectures across multiple datasets, employing a range of methods.

\subsection{Experimental Setup}
\subsubsection{Datasets} The datasets we use include the following: MNIST~\cite{lecun1998gradient}, F-MNIST~\cite{xiao2017fashion}, CIFAR-10~\cite{krizhevsky2009learning} and CIFAR-100~\cite{krizhevsky2009learning}, 
These datasets are widely used in the fields of computer vision and machine learning, serving as benchmarks for model evaluation and comparison.

\subsubsection{Training Model}
The training models implemented are from the VGG family~\cite{simonyan2014very}, mainly including VGG-13, VGG-16-Wider, VGG-19, and their variants, including VGG-14, VGG-15, VGG-17, VGG-18, and VGG-19-Wider. Among them, VGG-16-Wider is obtained by widening a certain layer of VGG-16, and VGG-19-Wider is generated similarly. The number of types of model architectures is set to $8$, in this case, $6$ clients will be trained using VGG-19, and the other $7$ models will be adopted by two clients each. 

\subsubsection{Baselines}
Three FL methods are adopted as baselines for comparison with our work: 
\begin{itemize}
    \item Standalone: In the Standalone learning approach, each client independently trains its model without sharing data or model updates with other clients.
    \item Clustered-FL~\cite{sattler2020clustered}: Clustered-FL~\cite{sattler2020clustered} enhances model training by clustering clients with similar structures, allowing for more precise and personalized model training within each cluster. Clients within a cluster share model updates, leading to improved model performance while effectively addressing data heterogeneity among clients. This approach balances personalization and generalization by facilitating collaborative learning among similar clients.
    \item FlexiFed~\cite{wang2023flexifed}: Clustered-Common is a key component of the FlexiFed method~\cite{wang2023flexifed}, designed specifically for FL scenarios with significant data heterogeneity~\cite{wang2023flexifed}. Clustered-Common clusters clients with similar data distributions and train a shared model for each cluster. These shared models are distributed within the cluster but not across different clusters, maintaining the balance between local personalization and global generalization.
\end{itemize}

\subsubsection{Other settings}
The number of clients $K$ is set to $20$, and the participating rate is set to 1, which means that all clients participate in each training round. Clients will use $20\%$ of their datasets in each round of training. Global training is carried out over $200$ rounds, with a learning rate of $lr=0.01$, a batch size of $64$, and the local training epoch $E$ is set to $10$ per round.

\subsection{Experimental Results}
To thoroughly evaluate the performance of FedADP under various conditions and to understand the impact of different factors, we designed a comprehensive series of experiments. These experiments encompassed multiple aspects: comparing the performance of FedADP in different datasets with baselines.

The results presented in Table 1 clearly indicate that FedADP consistently outperforms FlexiFed~\cite{wang2023flexifed}, Clustered-FL~\cite{sattler2020clustered}, and Standalone across all experimental conditions in terms of accuracy. This superiority highlights the effectiveness of FedADP in adapting to varying client architectures. Additionally, the findings illustrated in Fig. 4 demonstrate that both FedADP and FlexiFed~\cite{wang2023flexifed} exhibit significantly superior convergence rates compared to Clustered-FL~\cite{sattler2020clustered} and Standalone under all tested conditions. Notably, the difference in convergence speed between FedADP and FlexiFed~\cite{wang2023flexifed} is minimal, yet FedADP achieves a higher accuracy overall. These experimental results suggest that FedADP successfully strikes an effective balance between training performance and efficiency, showcasing a higher level of effectiveness in environments where clients have heterogeneous models. 

\section{CONCLUSION}
This paper introduces the FedADP approach, which addresses the issue of model heterogeneity by altering the model structure. Compared to state-of-the-art personalized FL approaches, FedADP demonstrates strong resilience to heterogeneity, maintaining high accuracy across diverse client environments, as evidenced by our experimental results.

\section{ACKNOWLEDGMENTS}
This work was partially supported by Natural Science Foundation
of Shandong (Shandong NSF No. ZR2021LZH006 and No. ZR2023QF083), Taishan Scholars
Program. Lei Liu is the corresponding author of this paper.

\balance
\bibliographystyle{unsrt}

\end{document}